\documentclass[krantz1,ChapterTOCs, oneside]{krantz} 

\usepackage{amssymb}
\usepackage{amsmath}
\usepackage{graphicx}
\usepackage{subfigure}
\usepackage{makeidx}
\usepackage{multicol}
\usepackage[export]{adjustbox}
\usepackage[justification=centering]{caption}
\usepackage{chngcntr}
\counterwithout{figure}{section}
\usepackage{float}
\usepackage[capitalise, noabbrev]{cleveref}

\begin{document}








\mainmatter

\renewcommand{\sectionmark}[1]{\markboth {}{}}
\renewcommand{\thesection}{\arabic{section}}


\chapter{PressureTransferNet: Human Attribute Guided Dynamic Ground Pressure Profile Transfer using 3D simulated Pressure Maps}


%
\vspace{-60pt}

\begin{multicols}{2}
\contributor{Lala Shakti Swarup Ray\footnote{lala\_shakti\_swarup.ray@dfki.de}}{DFKI Kaiserslautern}{}\\

\contributor{Vitor Fortes Rey}{DFKI and RPTU Kaiserslautern}{}\\

\contributor{Bo Zhou}{DFKI and RPTU Kaiserslautern}{}\\

\contributor{Sungho Suh}{DFKI and RPTU Kaiserslautern}{}\\

\contributor{Paul Lukowicz}{DFKI and RPTU Kaiserslautern}{}\\
\end{multicols}

\vspace{-30pt}
%
%
%
%
%
%
%
%
\section*{Abstract}
We propose PressureTransferNet, a novel method for Human Activity Recognition (HAR) using ground pressure information. Our approach generates body-specific dynamic ground pressure profiles for specific activities by leveraging existing pressure data from different individuals. PressureTransferNet is an encoder-decoder model taking a source pressure map and a target human attribute vector as inputs, producing a new pressure map reflecting the target attribute. To train the model, we use a sensor simulation to create a diverse dataset with various human attributes and pressure profiles. Evaluation on a real-world dataset shows its effectiveness in accurately transferring human attributes to ground pressure profiles across different scenarios. We visually confirm the fidelity of the synthesized pressure shapes using a physics-based deep learning model and achieve a binary R-square value of 0.79 on areas with ground contact. Validation through classification with F1 score (0.911$\pm$0.015) on physical pressure mat data demonstrates the correctness of the synthesized pressure maps, making our method valuable for data augmentation, denoising, sensor simulation, and anomaly detection. Applications span sports science, rehabilitation, and bio-mechanics, contributing to the development of HAR systems.
\section{Introduction}
\label{section:Introduction}
Human Activity Recognition (HAR) is a fundamental task that automatically identifies and classifies human activities based on input data from various sensors. HAR has gained significant attention in recent years due to its wide range of applications, including video surveillance \cite{khan2020human,nasir2022harednet}, healthcare monitoring \cite{javeed2023deep,ghadi2022intelligent}, human-computer interaction \cite{gong2022mmg,malibari2022quantum}, and sports analytics \cite{host2022overview,mekruksavanich2022sport}. Traditional HAR approaches primarily rely on visual data, such as video or image sequences, to recognize activities \cite{pareek2021survey}. However, solely relying on visual information can pose challenges in scenarios with limited or ambiguous visual cues.

To overcome these challenges, researchers have explored using additional sensor modalities to enhance HAR systems like accelerometers \cite{ayman2019efficient,zhang2022if,lim2021deep}, EMG \cite{nurhanim2021emg}, capacitive sensors \cite{zhou2023mocapose}, multi modal systems \cite{yadav2021review,muaaz2020wiwehar, adachi2022using} etc. One such promising sensor modality is ground pressure sensing \cite{antwi2020construction,ma2020adaptive}, which provides valuable information about the forces exerted by human activities on the ground. Ground pressure sensors can be embedded in flooring surfaces, footwear, or specialized sensor-equipped platforms to capture the spatiotemporal distribution of forces during various activities. By incorporating ground pressure data into HAR systems, it is possible to enhance activity recognition accuracy and robustness, especially in situations where visual cues alone may be insufficient or unreliable.
The number of datasets that includes ground pressure measurements is limited \cite{dos2017data}. The scarcity of this data can be attributed to various factors. The process of collecting ground pressure data is a time-consuming and labor-intensive task. Moreover, the terrain and environmental conditions significantly impact the data collection process, further complicating the efforts.
Creating a comprehensive dataset through physical measurements alone is a daunting challenge. As an alternative, a combination of physics simulation and deep learning models can be employed to generate ground pressure data \cite{ray2023pressim}. However, even simulations require expertise and substantial tuning to mimic real-world scenarios accurately. Developers must carefully calibrate simulation models based on empirical data to ensure the generated results align with actual observations. This tuning process demands considerable time, effort, and computational resources.
The limited availability of ground pressure datasets poses significant obstacles for various industries and research fields that rely on this information. Therefore, efforts should be directed toward expanding the existing datasets, employing advanced measurement techniques, and improving simulation methodologies to bridge the gap and provide reliable ground pressure information for practical applications.
Style transfer techniques have proven effective in augmenting and transforming visual data for various computer vision tasks, such as image synthesis \cite{an2021artflow}, text generation \cite{jin2022deep}, voice synthesis \cite{qian2019autovc} and motion translation \cite{aberman2020unpaired,smith2019efficient}. Style transfer involves extracting and transferring one modality's style or characteristics onto another while preserving the content. 
By leveraging style transfer techniques, it becomes possible to learn to map ground pressure characteristics into human attributes, enabling the translation of ground pressure sensor signals from one representation to another.

\begin{figure}[!t]
  \centering
  \includegraphics[width=\linewidth]{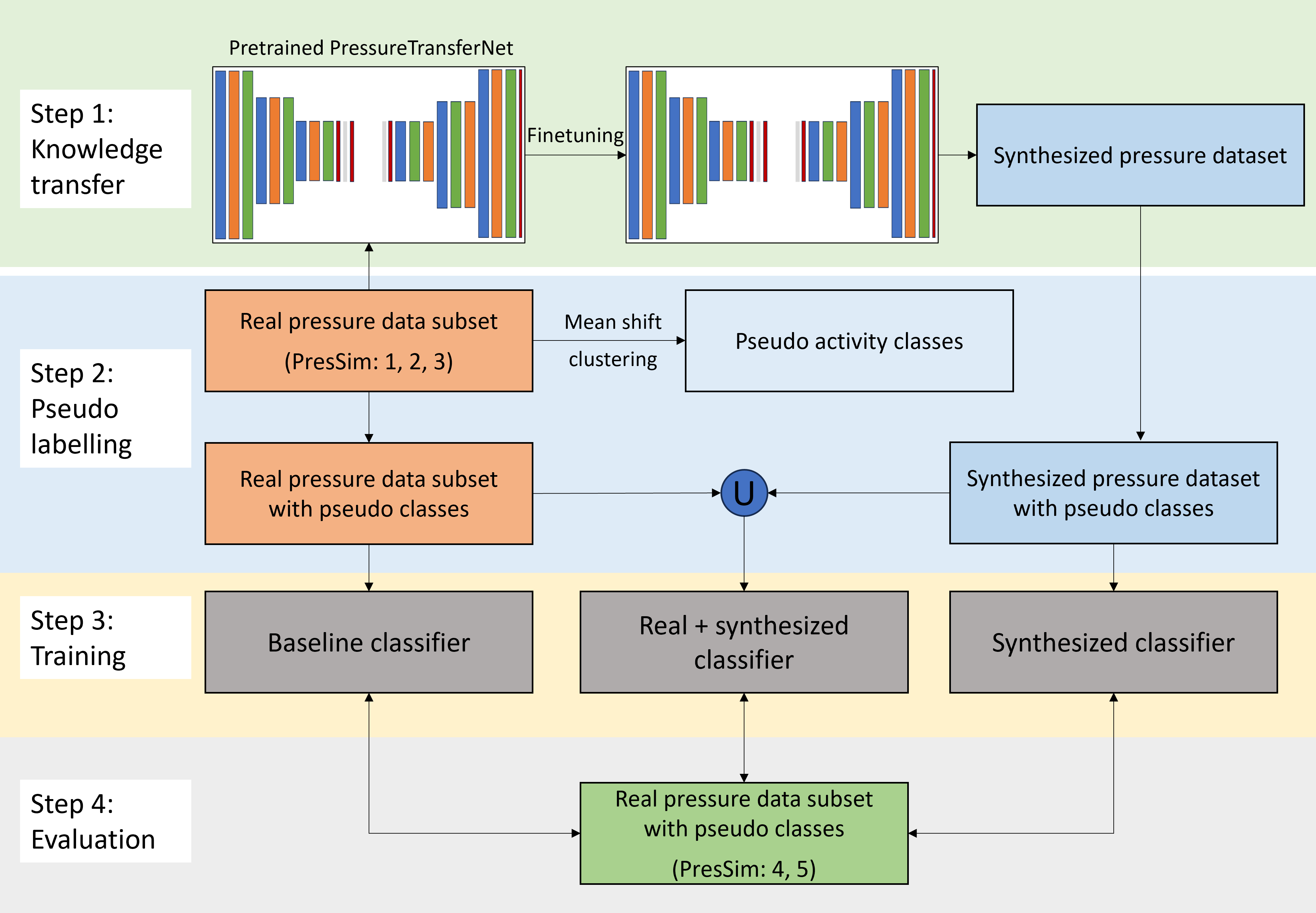}
  \caption{Validation of PressureTransferNet over real pressure data using unsupervised activity classification.}
  \label{fig:val}
\end{figure}

This paper proposes a novel approach combining ground pressure sensor data with style transfer techniques for enhancing HAR. 
Our method aims to find the correlation between sensor signal representations and body attributes and use that to synthesize the pressure of corresponding human activities for a specific body onto visual frames. 
By leveraging the complementary information provided by ground pressure sensing and visual data, our approach can improve the accuracy and robustness of HAR systems in challenging real-world scenarios.
Our main contributions include the following:
\begin{itemize}
    \item Framework to simulate sensor signals from dynamic poses using a combination of 3D simulation and deep learning, exemplified here by simulated pressure sensor dataset derived from TotalCapture dataset having four motion sequences performed by four different subjects. We also re-targeted the above motions on eight male and eight female SMPL bodies having different builds.
    \item PressureTransferNet: a U-net-inspired Encoder-decoder neural network to synthesize style transferred signals for existing pressure maps and human attributes, analysis of the synthesized augmented pressure sensor on simulated ground truth of users of the target dataset motions not used while training the model as well validation using real pressure data as depicted in \cref{fig:val}. 
\end{itemize}
\begin{figure}[!t]
  \centering
  \includegraphics[width=\linewidth]{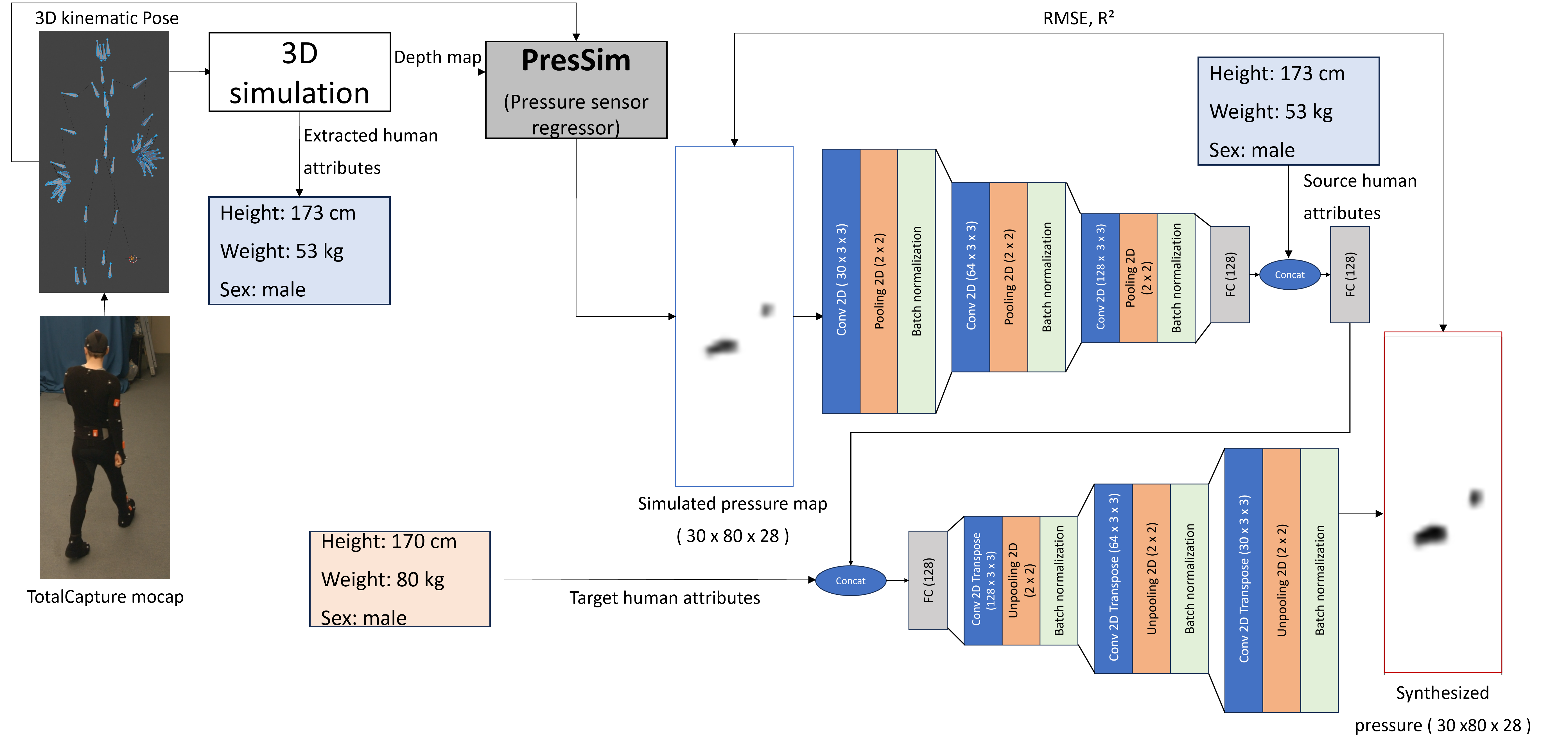}
  \caption{Overall pipeline of PressureTransferNet depicting data generation using PresSim and training.}
  \label{fig:pipeline}
\end{figure}

The remainder of this paper is organized as follows: \cref{sec:reclinkage} provides an overview of related work in HAR, ground pressure sensing, and style transfer. \cref{sec:3} describes the methodology and the proposed sensor style transfer approach. In \cref{sec: res}, we present experimental results and performance evaluation on benchmark datasets. Finally, \cref{sec: 5} concludes the paper and discusses future directions for research in the field of HAR with sensor style transfer.
\section{Related Work}
\label{sec:reclinkage}
In pressure sensor simulation, a virtual environment is created to replicate the behavior and characteristics of a real pressure sensor. Using simulated pressure values, the algorithm generates corresponding output signals that mimic the sensor's behavior in real-world conditions. These simulated values are then translated to real pressure values using a neural network trained on parallel simulated and real pressure maps. This virtual simulation allows for the analysis of different pressure levels, environmental factors, and dynamic changes, providing valuable insights for system design, optimization, and performance evaluation. 
There are various works where pressure data is being simulated from other modalities such as RGB, depth images \cite{davoodnia2021bed,clever2022bodypressure} or 3D poses \cite{scott2020kinematics,ray2023pressim}.
One notable work is PresSim \cite{ray2023pressim}, where synthetic pressure sensor data is created from 3D volumetric poses using neural networks and 3D simulations. This synthesis involves a 4-stage process of multimodal data acquisition, pose and shape estimation using SMPL models (skinned multi-person linear model) \cite{loper2015smpl}, physics simulation, and 3D regressor DL (deep learning) model to map the simulated data to real pressure maps. Validation experiments using a monocular camera and a pressure-sensing fitness mat demonstrated a high level of agreement between the synthesized pressure maps and the pressure sensor's ground truth, with a $R^2$ value of 0.811.
We build upon the sensor simulation from PresSim to generate synthetic pressure sensor data variations using our proposed PressureTransferNet. It augments PresSim simulations by incorporating human attributes with style transfer in our Encoder-decoder architecture.
The overall pipeline is depicted in \cref{fig:pipeline}.

\section{Proposed Method}
\label{sec:3}
Since different subjects cannot repeat the exact motion sequence, real-world data cannot be used to train our proposed style transfer model. We bridge this gap by relying on sensor simulation, with our work being based on the assumption that PresSim synthesizes accurate pressure maps.
Our workflow involves a two-stage process of synthetic data simulation using PresSim and SMPL-X \cite{pavlakos2019expressive} blender plugin. We generate body-specific pressure data from input pressure maps and body attributes using DL-based PressureTransferNet.

\subsection{Data Generation}
To provide ground truth pressure maps for TotalCapture,
we used a combination of PresSim, Blender3D and the SMPL-X blender plugin to generate our dataset.
Our data are derived from the TotalCapture dataset.
We consider all 16 unique motions categorized into four categories: walking, exercising, freestyle, and acting.
walking1,rom1,freestyle1, and acting1 from subject 1 (s1), subject 2 (s2), subject 3 (s3) and subject 4 (s4) are considered while generating our dataset.
The resulting data contains 308400 frames of synthesized ground pressure data along with extracted three human attributes gender, weight, and height of the virtual human SMPL used for synthesizing the dataset.
Detailed data statistics are provided in \cref{table1}.

\begin{table}
\caption{ Data statistics: Source MoCap file from TotalCaptue dataset and a total number of simulated frames for each motion category.}
\label{table1}
  \begin{center}
    {\small{
\begin{tabular}{p{6cm}|p{2cm}} 
Motion & Total frames \\
\hline
walking1(s1, s2, s3, s4) & 73360 \\
rom1(s1, s2, s3, s4) & 97760 \\
freestyle1(s1, s2, s3, s4)  & 53760 \\
acting1(s1, s2, s3, s4) & 83520 \\
\hline
Total & 308400 \\
\end{tabular}
}}
\end{center}

\end{table}

For our synthetic augmented data based on human attributes,
we considered 20 human models, ten males with a height range of 175$\pm$ 15 cm and a weight range of 75$\pm$10 kg and ten females with a height range of 165$\pm$ 15 cm and a weight range 65$\pm$ 10 kg randomly distributed using the Gaussian distribution.
The models are generated inside Blender along with the 24-joint skeleton, which is used as a base to transfer the pose sequence from TotalCapture \cite{joo2018total} dataset to Blender using AMASS \cite{mahmood2019amass} and motion re-targeting.

\begin{figure}[!t]
  \centering
  \includegraphics[width=\linewidth]{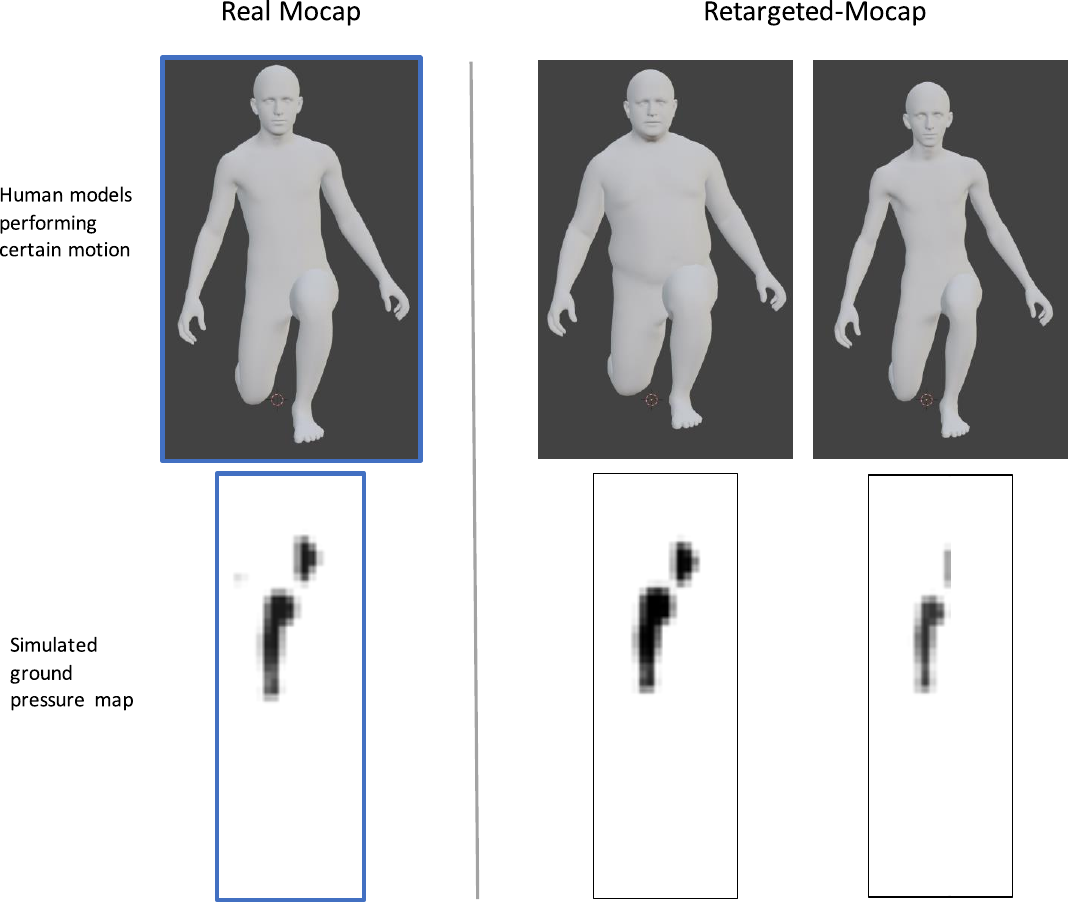}
  \caption{Figure depicting simulated ground pressure maps from certain activity performed by the real motion-captured model and virtual models with re-targeted motions inside Blender3D. } 
  \label{fig:sim}
\end{figure}

The PresSim pipeline was used as a basis for simulating our pressure sensor maps, with a view-port added to the scene and positioned just below the human model to capture the best angle for estimating ground pressure. Camera settings such as focal length, aperture, depth pass, and focus distance are configured to consider only the lowest points. 80*28 OpenEXR image files are rendered as outputs, which will be saved to the specified location. We added custom compositing nodes that, in turn, takes the depth map sequences and adds a multiplier equivalent to the model weight only to all pixels with value more than 0. 
The simulated intermediate data looked almost identical to the intermediate data generated by PresSim while preserving a lot of complex simulations and computations.
We then used the pre-trained PresSim regressor based neural network with our simulated depth maps and the actual 3D joint position sequences to generate body-specific pressure sensor maps having 80*28 sensor nodes, as depicted in \cref{fig:sim}.
Our pressure map variations thus add parallel motion sequences where different human models perform the same motion sets as in the original MoCap data.

\subsection{Network Architecture}
We used a U-Net-inspired Encoder-decoder model to train PressureTransferNet. The U-Net architecture is commonly used for image segmentation tasks and allows for the preservation of spatial information. Since pressure data can be easily represented in image space, the U-Net-inspired architecture helps to generate accurate and detailed pressure sensor maps by effectively capturing the relevant features and spatial dependencies (intrinsic motions) in the input data. The encoder with 11 layers takes 30 consecutive frames of pressure sensor data of size 80 $\times$ 28 to generate an intermediate latent vector using three blocks of 2D convolution, batch normalization (BN), and pooling. This is followed by one fully connected (FC) layer to generate the intermediate representation of size 128. Three source human attributes (height, weight, and sex) are concatenated with the intermediate latent vector and passed through another FC layer to generate our final latent representation of size 128. The main hypothesis behind doing it is first to extract the relevant features that define the underlying motion irrespective of the body type used for the pressure synthesis and, in the end, concatenate the subject information with the latent information to give input to the decoder.

The decoder receives this representation concatenated with the target human attributes of the desired new body shape. Since it wants to generate a body-specific pressure map sequence, it makes sense to have all the information, i.e., underlying motion representation, source body information, and target body information, as input to the decoder. Its architecture is a mirrored version of the encoder using three blocks of Conv2DTranspose followed by UnPooling and BN layers. It outputs the resulting pressure sensor map for 30 frames, each with 80*28 pressure nodes. 

We used content loss to train our neural network, which is a common type of loss used while working on style transfer networks. It is defined as:
\[
\text{{Content Loss}} = \frac{1}{N} \sum_{i=1}^{N} \| F_i^{\text{{input}}} - F_i^{\text{{generated}}} \|^2
\]

Where \(\text{{Content Loss}}\) represents the content-related difference between the input and generated pressure maps. \(F_i^{\text{{input}}}\) and \(F_i^{\text{{generated}}}\) denote the feature maps of the input and generated pressure sequences, respectively. \(N\) represents the total number of feature maps.

\section{Evaluation}
\label{sec: res}
We use as instances sliding windows of motion with a size of 30 frames with a one-frame stride. Our experiments have 20 subjects (4 from the original MoCap and 16 re-targeted variations). The dataset is divided into train, test, and validation. For training, we use the simulated sensor data from 8 male and female subjects. At the same time, the test set consists of the motions of the original two male and female participants.
The training set is randomly divided into train and validation using a 9:1 ratio.
The neural network was trained and tested on a workstation with a GPU accelerator (NVidia GeForce RTX 3080) with a batch size of 1028, Adam optimizer with an initial learning rate of 0.01 which is reduced by half every 20 epochs.
The model is trained for 500 epochs using 50 epochs of patience for early stopping.

\subsection{Evaluation Metrics}
We used two metrics to evaluate our model on both simulated ground truth on seen and unseen motion sequences (kept out of the training or test set, see \cref{table2}).
For evaluating PressureTransferNet, we used both the root mean square error (RMSE) and binary $R^2$ scores.
RMSE is defined as:
\[
\text{{RMSE}} = \sqrt{\frac{1}{N} \sum_{t=1}^{N} (P_t^{\text{{synthesized}}} - P_t^{\text{{ground truth}}})^2}
\]

Where \(N\) is the total number of time steps or frames in the pressure map sequence. \(P_t^{\text{{synthesized}}}\) and \(P_t^{\text{{ground truth}}}\) denote the pressure maps at time step \(t\) in the resulting and input sequences, respectively.
It provides a measure of the average difference between the resulting and input pressure map sequences, with lower values indicating better similarity or accuracy.
To compare the synthesized pressure map shapes with that of simulated ones, we used binary $R^2$, which can be represented as follows:
\[
\text{{Binary }} R^2 = 1 - \frac{{\sum_{t=1}^{N}(P_t^{\text{{synthesized}}} - P_t^{\text{{ground truth}}})^2}}{{\sum_{t=1}^{N}(P_t^{\text{{ground truth}}} - \bar{P}^{\text{{ground truth}}})^2}}
\]

Where \(\bar{P}^{\text{{ground truth}}}\) represents the average value of the input pressure map sequence. A value closer to 1 indicates a better match between the resulting and input pressure map sequences. The pressure maps are considered binary variables for this equation, meaning that any nonzero value is treated as one, and zero values remain zero.
\subsection{Results}
\begin{figure}[!t]
  \centering
  \includegraphics[width=\linewidth]{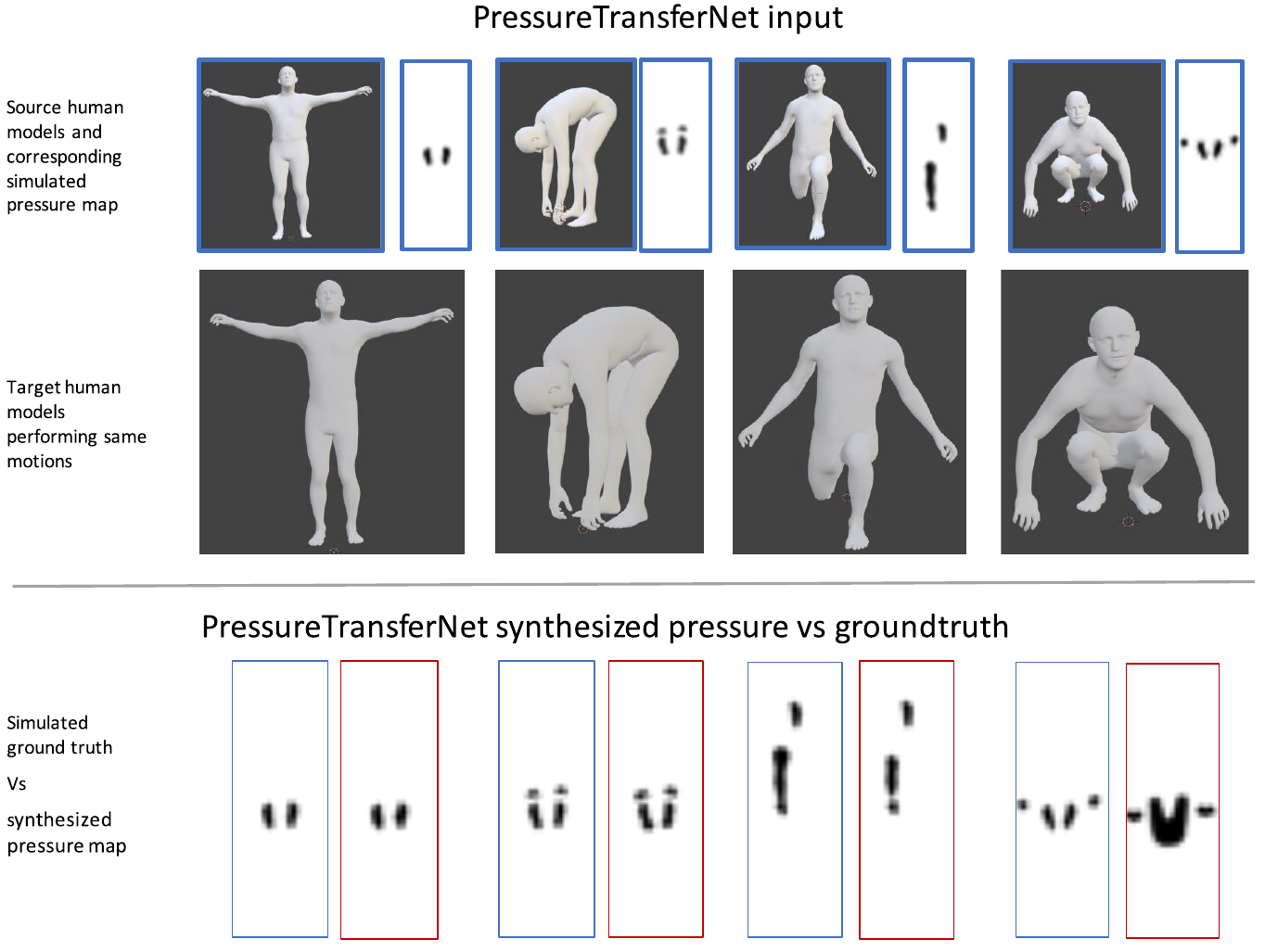}
  \caption{Qualitative comparison of simulated pressure ground truth and synthesized pressure map using PressureTransferNet.} 
  \label{fig:res}
\end{figure}
For qualitative evaluation of the simulated ground truth, we compared our synthesized pressure maps with the simulated pressure maps used for training PressureTransferNet.
Figure \ref{fig:res} depicts different motion sequences, simulated pressure ground truth, and Synthesized pressure map using PressureTransferNet, respectively.
For quantitative evaluation, we divided the motion sequences into two categories: seen and unseen.
Seen motion sequences refer to the motion sequences included in the TotalCaptureDataset, which were used to create the synthetic dataset for training PressureTransferNet. These sequences belong to the subjects s1, s2, s3, or s4.
Unseen motion sequences, on the other hand, are also derived from the TotalCaptureDataset. However, these sequences belonged to the subject s5 and were not used during the creation of the synthetic dataset to train PressureTransferNet. Despite belonging to the same four motion classes as the seen sequences, they were not part of the training process.
The purpose of using both seen and unseen motion sequences is to evaluate the performance of PressureTransferNet on familiar and unfamiliar data. Seen sequences allow us to assess how well the model generalizes to the data it was trained on, while unseen sequences indicate the model's ability to handle novel or previously unseen motion patterns.
For the seen motion sequences, we achieved an average RMSE of 8.88 and an average binary $R^2$ 0.79, which depicts a high correlation between synthesized pressure sequences and ground truth. For unseen motion sequences sharing the same motion categories, the average RMSE is 12.82 and the average binary $R^2$ 0.70, which is worse than for the seen motion sequences but still acceptable.
The individual and average score for each motion category for unseen and seen motion is given in \cref{table2}.

\begin{table}
\caption{ Binary $R^2$ and RMSE for all motion categories for TotalCapture motion sequences used to create a dataset, as well as motion sequence not used while creating a dataset for PressureTransferNet.}
\label{table2}
  \begin{center}
    {\small{
\begin{tabular}{p{6cm}|p{2cm}p{2cm}} 

Motion Category & RMSE & Binary $R^2$ \\
\hline
Seen motions&&\\
\hline
walking1(s1, s2, s3, s4) & 8.72 & 0.81 \\
rom1(s1, s2, s3, s4) & 9.21 & 0.78 \\
freestyle1(s1, s2, s3, s4) & 8.93 & 0.80 \\
acting1(s1, s2, s3, s4) & 9.67 & 0.76 \\
Average score & 8.88 & 0.79 \\
\hline
Unseen motions&&\\
\hline
walking2(s1, s2, s3, s4) & 12.46 & 0.71 \\
rom2(s1, s2, s3, s4) & 12.89 & 0.69 \\
freestyle2(s1, s2, s3, s4) & 13.22 & 0.70 \\
acting2(s1, s2, s3, s4) & 12.73 & 0.69 \\
Average score & 12.82 & 0.70 \\
\end{tabular}
}}
\end{center}
\end{table}

To check the correctness of PressureTransferNet compared to real ground pressure data, we fine-tuned our model by training it with a subset of the PresSim dataset containing the first 20-minute sequence for each subject 1,2 and 3 12832 frames along with 38496 frames of simulated pressure maps.
Because of the lack of activity labels, we generated pseudo labels based on the mean shift algorithm.
A classifier is trained on the real pressure maps, synthesized pressure maps from PressureTransferNet as well as the combination of real and synthesized pressure maps from subjects 1,2 and 3 using the pseudo labels and evaluated using the real pressure data from subjects 4 and 5 along with the generated pseudo labels.
The result showed improvement in f1 score when trained on a combination of real and PressureTransferNet generated data as depicted in \cref{table3}, and in term proves that the synthesized pressure maps are of a good enough quality to train a HAR system.
\begin{table}
\caption{ Comparison of macro F1 score for Unsupervised classification model trained on real pressure data, synthetic pressure data generated by PressureTransferNet, and evaluated on real pressure data for ten iterations.}
\label{table3}
  \begin{center}
    {\small{
\begin{tabular}{p{4cm}|p{2cm}p{2cm}p{2cm}}\\
Training dataset source & Real & Synthetic & Real+Synthetic \\
\hline
Macro F1 score & 0.879$\pm$0.014 & 0.803$\pm$0.02 & 0.911$\pm$0.015 \\
\end{tabular}
}}
\end{center}
\end{table}
\section{Conclusion}
\label{sec: 5}

In this work, we have proposed PressureTransferNet,  a novel approach that combines ground pressure sensor data with style transfer techniques to generate human attribute-guided pressure maps, generating new variations for existing motions. We have validated our method both qualitatively and quantitatively, with reasonable results that point to the potential of the technique for data augmentation in HAR applications. 

In future work, we can use real pressure data and manually labeled activity classes as a starting point for our variations and input for training PressureTransferNet and evaluating HAR based models.
Another areas to explore are the attributes used to guide style transfer. We can also investigate using SMPL shape parameters for style transfer instead of our three physical parameters, which provide only partial information on body shape compared to the full SMPL description.
The same workflow can also be applied to other sensor modalities, such as IMU or EMG, to create style transfer models. 
\section{Acknowledgements}
\label{sec: 6}
The research reported in this paper was supported by the BMBF (German Federal Ministry of Education and Research) in the project VidGenSense (01IW21003). It was also funded by Carl-Zeiss Stiftung under the Sustainable Embedded AI project (P2021-02-009).

\bibliographystyle{frontmatter/spmpsci.bst}
\bibliography{Authors/bibtex}

\end{document}